\documentclass[10pt,twocolumn,letterpaper]{article}

\usepackage{cvpr}              % To produce the CAMERA-READY version
\usepackage{makecell}
\usepackage{multirow}
\usepackage{graphicx}

\usepackage[table]{xcolor}
\usepackage{xcolor}
\usepackage{appendix}
\definecolor{mr_color}{RGB}{255, 128, 0}

\usepackage{amsmath,amsfonts,bm}

\def\eqref#1{equation~(\ref{#1})}
\def\1{\bm{1}}

\DeclareMathAlphabet{\mathsfit}{\encodingdefault}{\sfdefault}{m}{sl}
\SetMathAlphabet{\mathsfit}{bold}{\encodingdefault}{\sfdefault}{bx}{n}

\definecolor{cvprblue}{rgb}{0.21,0.49,0.74}

\usepackage[pagebackref,breaklinks,colorlinks,citecolor=cvprblue]{hyperref}

\def\paperID{5582} % *** Enter the Paper ID here
\def\confName{CVPR}
\def\confYear{2024}

\title{RMT: Retentive Networks Meet Vision Transformers}

\author{%
  Qihang Fan $^{1, 2}$, Huaibo Huang$^{1}$, Mingrui Chen$^{1, 2}$, Hongmin Liu$^{3}$, Ran He$^{1, 2}$\thanks{Ran He is the corresponding author.}\\
  $^1$MAIS \& CRIPAC, Institute of Automation, Chinese Academy of Sciences, Beijing, China\\
  $^2$School of Artificial Intelligence, University of Chinese Academy of Sciences, Beijing, China\\
  $^3$University of Science and Technology Beijing, Beijing, China\\
  \texttt{fanqihang.159@gmail.com, huaibo.huang@cripac.ia.ac.cn,}\\
  \texttt{charmier@hust.edu.cn, hmliu\_82@163.com, rhe@nlpr.ia.ac.cn}\\
}

\begin{document}
\maketitle
\begin{abstract}

Vision Transformer (ViT) has gained increasing attention in the computer vision community in recent years. However, the core component of ViT, Self-Attention, lacks explicit spatial priors and bears a quadratic computational complexity, thereby constraining the applicability of ViT. To alleviate these issues, we draw inspiration from the recent Retentive Network (RetNet) in the field of NLP, and propose RMT, a strong vision backbone with explicit spatial prior for general purposes. Specifically, we extend the RetNet's temporal decay mechanism to the spatial domain, and propose a spatial decay matrix based on the Manhattan distance to introduce the explicit spatial prior to Self-Attention. Additionally, an attention decomposition form that adeptly adapts to explicit spatial prior is proposed, aiming to reduce the computational burden of modeling global information without disrupting the spatial decay matrix. Based on the spatial decay matrix and the attention decomposition form, we can flexibly integrate explicit spatial prior into the vision backbone with linear complexity. Extensive experiments demonstrate that RMT exhibits exceptional performance across various vision tasks. Specifically, without extra training data, RMT achieves \textbf{84.8\%} and \textbf{86.1\%} top-1 acc on ImageNet-1k with \textbf{27M/4.5GFLOPs} and \textbf{96M/18.2GFLOPs}. For downstream tasks, RMT achieves \textbf{54.5} box AP and \textbf{47.2} mask AP on the COCO detection task, and \textbf{52.8} mIoU on the ADE20K semantic segmentation task. Code is available at \url{https://github.com/qhfan/RMT}

\end{abstract}    
\section{Introduction}
\label{sec:intro}

\begin{figure}[t]
\captionsetup{font=small}%
\scriptsize
\begin{center}
\includegraphics[width=0.95\linewidth]{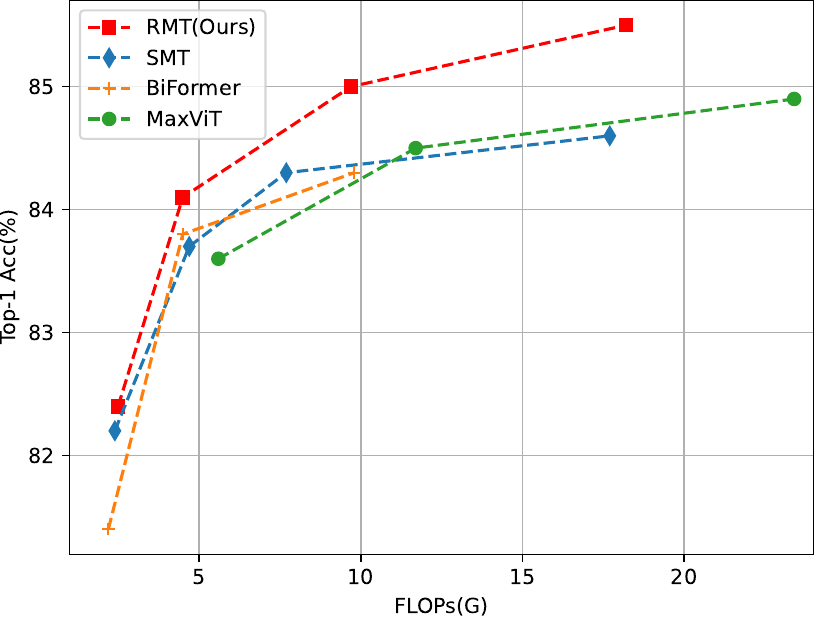}\label{fig:param_psnr}%
\hspace{-40mm}\resizebox{.45\columnwidth}{!}{
\begin{tabular}[b]{lcc}
    \footnotesize
	Model & \#Params & Top1 Acc.\\
    \hline
	MaxViT-T~\cite{mvitv2} & 31M & 83.6  \\ 
	SMT-S~\cite{SMT} & 20M & 83.7  \\ 
        BiFormer-S~\cite{biformer} & 26M & 83.8 \\
        \textbf{RMT-S (Ours)} & 27M & \textbf{84.1} \\ 
        \textbf{RMT-S* (Ours)} & 27M & \textbf{84.8} \\ 
    \hline
        
        BiFormer-B~\cite{biformer} & 57M & 84.3 \\
        MaxViT-S~\cite{mpvit} & 69M & 84.5 \\
        \textbf{RMT-B (Ours)} & 54M & \textbf{85.0} \\ 
        \textbf{RMT-B* (Ours)} & 55M & \textbf{85.6} \\ 
    \hline
        SMT-L~\cite{SMT} & 81M & 84.6 \\ 
	MaxViT-B~\cite{maxvit} & 120M & 84.9 \\ 
    \textbf{RMT-L (Ours)} & 95M & \textbf{85.5} \\ 
    \textbf{RMT-L* (Ours)} & 96M & \textbf{86.1} \\ 
	\hline
	\multicolumn{3}{c}{\vspace{10mm}}
\end{tabular}
}
\vspace{-0.3cm}
%\caption{\#Params vs. Top-1 accuracy on ImageNet.}
\caption{FLOPs v.s. Top-1 accuracy on ImageNet-1K  with $224\times 224$ resolution. ``*" indicates the model trained with token labeling~\cite{tokenlabel}.}
\label{fig:flop_vs_acc}
\end{center}
\vspace{-0.8cm}
\end{figure}

Vision Transformer (ViT)~\cite{vit} is an excellent visual architecture highly favored by researchers. However, as the core module of ViT, Self-Attention's inherent structure lacking explicit spatial priors. Besides, the quadratic complexity of Self-Attention leads to significant computational costs when modeling global information. These issues limit the application of ViT.

Many works have previously attempted to alleviate these issues~\cite{uniformer, SwinTransformer, CaiT, cloformer, cvt, cmt, LVT}. For example, in Swin Transformer~\cite{SwinTransformer}, the authors partition the tokens used for self-attention by applying windowing operations. This operation not only reduces the computational cost of self-attention but also introduces spatial priors to the model through the use of windows and relative position encoding. In addition to it, NAT~\cite{NAT} changes the receptive field of Self-Attention to match the shape of convolution, reducing computational costs while also enabling the model to perceive spatial priors through the shape of its receptive field. 

\begin{figure*}[t]
    \centering
    \includegraphics[width=0.99\linewidth]{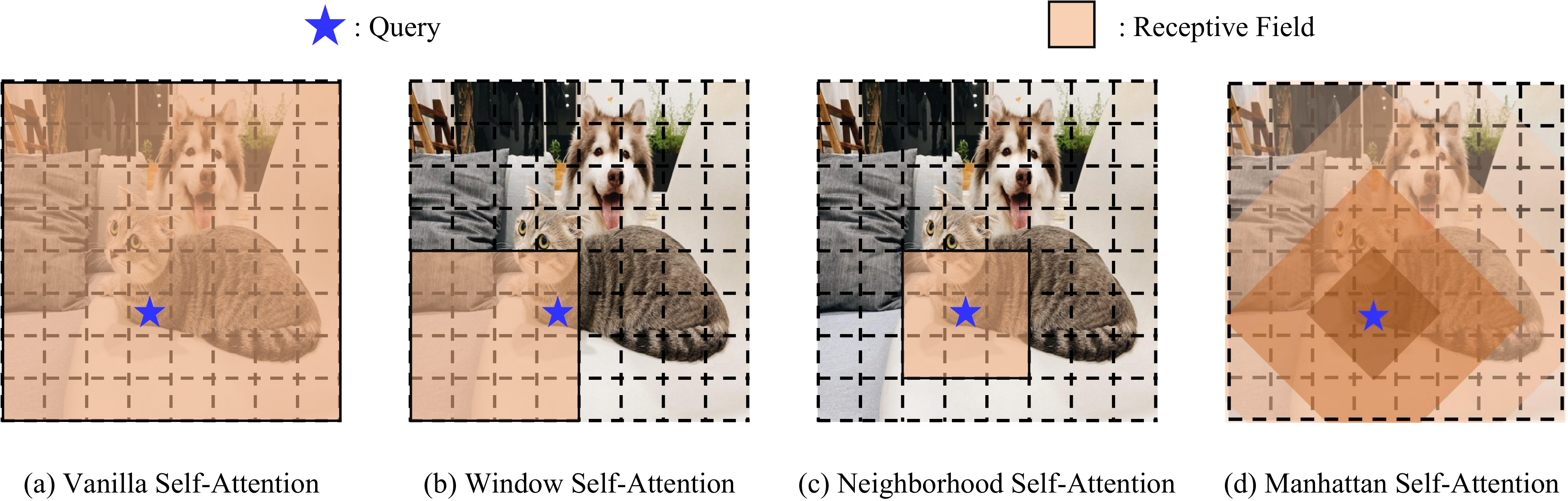}
    \caption{Comparison among different Self-Attention mechanisms. In MaSA, darker colors represent smaller spatial decay rates, while lighter colors represent larger ones. The spatial decay rates that change with distance provide the model with rich spatial priors.}
    \vspace{-0mm}
    \label{fig:intro_resa}
\end{figure*}

Different from previous methods, we draw inspiration from the recently successful Retentive Network (RetNet)~\cite{retnet} in the field of NLP. RetNet utilizes a distance-dependent temporal decay matrix to provide explicit temporal prior for one-dimensional and unidirectional text data. ALiBi~\cite{alibi}, prior to RetNet, also applied a similar approach and succeeded in NLP tasks. We extend this temporal decay matrix to the spatial domain, developing a two-dimensional bidirectional spatial decay matrix based on the Manhattan distance among tokens. In our space decay matrix, for a target token, the farther the surrounding tokens are, the greater the degree of decay in their attention scores. This property allows the target token to perceive global information while simultaneously assigning different levels of attention to tokens at varying distances. We introduce explicit spatial prior to the vision backbone using this spatial decay matrix. We name this Self-Attention mechanism, which is inspired by RetNet and incorporates the Manhattan distance as the explicit spatial prior, as \textbf{Ma}nhattan \textbf{S}elf-\textbf{A}ttention (MaSA). 

Besides explicit spatial priors, another issue caused by global modeling with Self-Attention is the enormous computational burden. Previous sparse attention mechanisms~\cite{cswin, SwinTransformer, pvt, focal, biformer} and the way retention is decomposed in RetNet~\cite{retnet} mostly disrupt the spatial decay matrix, making them unsuitable for MaSA. In order to sparsely model global information without compromising the spatial decay matrix, we propose a method to decompose Self-Attention along both axes of the image. This decomposition method decomposes Self-Attention and the spatial decay matrix without any loss of prior information. The decomposed MaSA models global information with linear complexity and has the same receptive field shape as the original MaSA. We compare MaSA with other Self-Attention mechanisms in Fig.~\ref{fig:intro_resa}. It can be seen that our MaSA introduces richer spatial priors to the model than its counterparts.

Based on MaSA, we construct a powerful vision backbone called RMT. We demonstrate the effectiveness of the proposed method through extensive experiments. As shown in Fig.~\ref{fig:flop_vs_acc}, our RMT outperforms the state-of-the-art (SOTA) models on image classification tasks. Additionally, our model exhibits more prominent advantages compared to other models in tasks such as object detection, instance segmentation, and semantic segmentation. Our contributions can be summarized as follows:

\begin{itemize}
    \item We propose a spatial decay matrix based on Manhattan distance to augment Self-Attention, creating the Manhattan Self-Attention (MaSA) with an explicit spatial prior.

    \item We propose a decomposition form for MaSA, enabling linear complexity for global information modeling without disrupting the spatial decay matrix.

    \item Leveraging MaSA, we construct RMT, a powerful vision backbone for general purposes. RMT attains high top-1 accuracy on ImageNet-1k in image classification without extra training data, and excels in tasks like object detection, instance segmentation, and semantic segmentation.
    
\end{itemize}

\section{Related Work}
\label{sec:formatting}
\newcommand{\op}[1]{\operatorname{#1}}
\newcommand{\mbf}[1]{\mathbf{#1}}
\paragraph{Transformer.}Transformer architecture was firstly proposed in \cite{attention} to address the training limitation of recurrent model and then achieve massive success in many NLP tasks. By splitting the image into small, non-overlapped patches sequence, Vision Transformer (ViTs)~\cite{vit} also have attracted great attention and become widely used on vision tasks~\cite{wavevit, tnt, dat, dfvit, litv2, twins}. Unlike in the past, where RNNs and CNNs have respectively dominated the NLP and CV fields, the transformer architecture has shined through in various modalities and fields~\cite{unifiedio, mplug2, CLIP, ALIGN}. In the computer vision community, many studies are attempting to introduce spatial priors into ViT to reduce the data requirements for training~\cite{CPVT, deit, NAT}. At the same time, various sparse attention mechanisms have been proposed to reduce the computational cost of Self-Attention~\cite{pvt, pvtv2, cloformer, cvt}.

\vspace{-4mm}

\paragraph{Prior Knowledge in Transformer.} Numerous attempts have been made to incorporate prior knowledge into the Transformer model to enhance its performance. The original Transformers~\cite{vit, attention} use trigonometric position encoding to provide positional information for each token. In vision tasks, \cite{SwinTransformer} proposes the use of relative positional encoding as a replacement for the original absolute positional encoding. \cite{CPVT} points out that zero padding in convolutional layers could also provide positional awareness for the ViT, and this position encoding method is highly efficient. In many studies, Convolution in FFN~\cite{cmt, pvtv2, cloformer} has been employed for vision models to further enrich the positional information in the ViT. For NLP tasks, in the recent Retentive Network~\cite{retnet}, the temporal decay matrix has been introduced to provide the model with prior knowledge based on distance changes. Before RetNet, ALiBi~\cite{alibi} also uses a similar temporal decay matrix.

\section{Methodology}
\begin{figure*}[ht]
    \centering
    \includegraphics[width=0.95\linewidth]{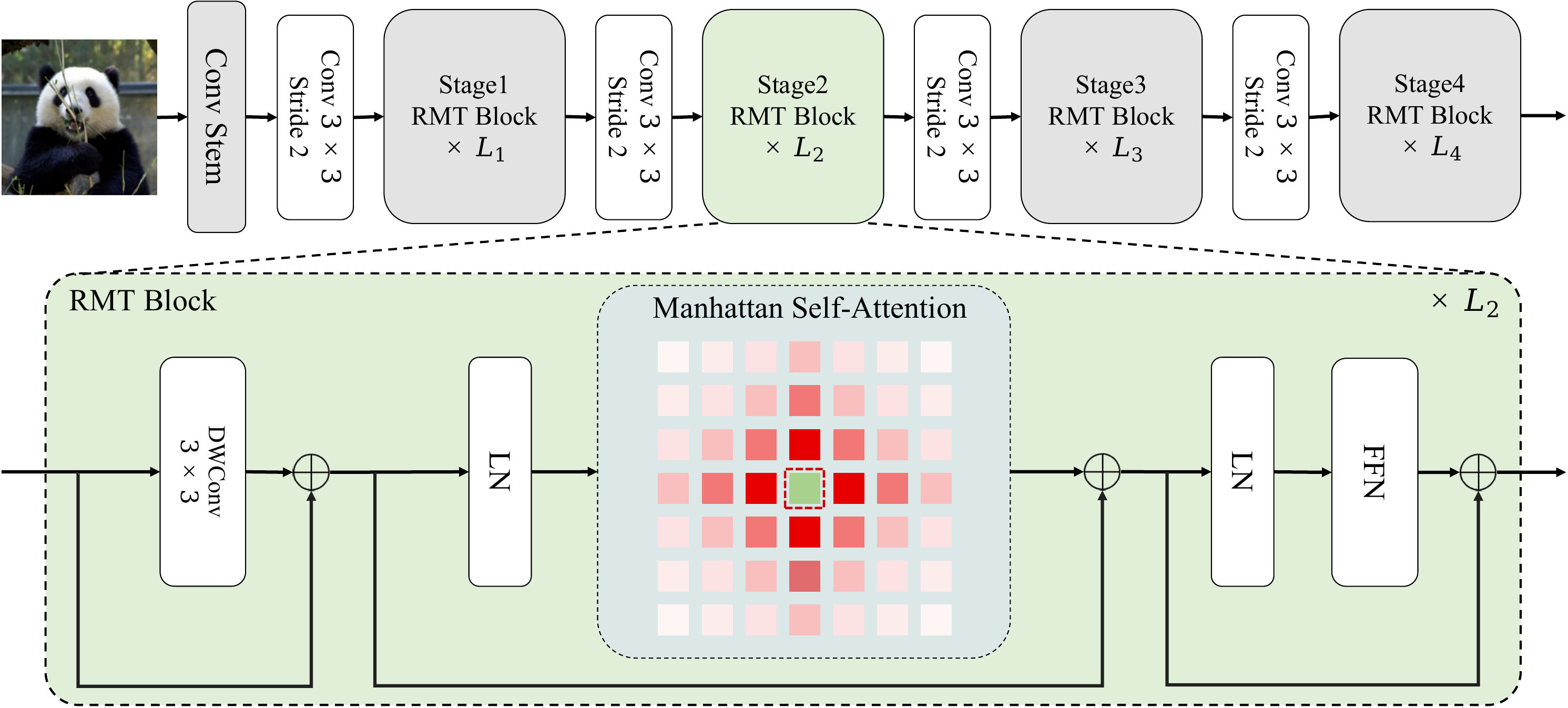}
    \caption{Overall architecture of RMT.}
    \label{fig:pipeline}
\end{figure*}

\subsection{Preliminary}
\paragraph{Temporal decay in RetNet.}Retentive Network (RetNet) is a powerful architecture for language models. This work proposes the retention mechanism for sequence modeling. Retention brings the temporal decay to the language model, which Transformers do not have. Retention firstly considers a sequence modeling problem in a recurrent manner. It can be written as Eq.~\ref{eq:ret:element}:
\begin{equation}
\label{eq:ret:element}
o_n = \sum_{m=1}^n \gamma^{n-m} (Q_n e^{in\theta})(K_m e^{im\theta})^\dag v_m
\end{equation}
For a parallel training process, Eq.~\ref{eq:ret:element} is expressed as:
\begin{equation}
\begin{aligned}
\label{eq:ret:parallel}
Q = (X W_Q) \odot \Theta ,&\quad K = (X W_K) \odot \overline{\Theta} ,\quad V = X W_V \\
\Theta_n = e^{in\theta},& \quad
D_{nm} =
\left\{
\begin{aligned}
& \gamma^{n-m}, &n\ge m \\
& 0, &n < m \\
\end{aligned}
\right.
\\
\mathrm{Rete}&\mathrm{ntion}(X) = (Q K^\intercal \odot D)V
\end{aligned}
\end{equation}
where $\overline{\Theta}$ is the complex conjugate of $\Theta$, and $D \in \mathbb{R}^{|x|\times |x|}$ contains both causal masking and exponential decay, which symbolizes the relative distance in one-dimensional sequence and brings the explicit temporal prior to text data.

\subsection{Manhattan Self-Attention}

Starting from the retention in RetNet, we evolve it into Manhattan Self-Attention (MaSA). Within MaSA, we transform the unidirectional and one-dimensional temporal decay observed in retention into bidirectional and two-dimensional spatial decay. This spatial decay introduces an explicit spatial prior linked to Manhattan distance into the vision backbone. Additionally, we devise a straightforward approach to concurrently decompose the Self-Attention and spatial decay matrix along the two axes of the image.
%Starting from the retention in RetNet, we develop the Manhattan Self-Attention. In MaSA, we modify the unidirectional and one-dimensional temporal decay in retention to the bidirectional and two-dimensional spatial decay. The spatial decay brings the explicit spatial prior associated with Manhattan distance to the vision backbone. Besides, we design a simple method to simultaneously decompose the Self-Attention and spatial decay matrix to the two axes of the image. 

\paragraph{From Unidirectional to Bidirectional Decay:} In RetNet, retention is unidirectional due to the causal nature of text data, allowing each token to attend only to preceding tokens and not those following it. This characteristic is ill-suited for tasks lacking causal properties, such as image recognition. Hence, we initially broaden the retention to a bidirectional form, expressed as Eq.~\ref{eq:biret:para}:
\begin{equation}
    \label{eq:biret:para}
    \begin{aligned}
        \mathrm{BiRetention}(X) &= (Q K^\intercal \odot D^{Bi})V 
        \\
        D_{nm}^{Bi} &= \gamma^{|n-m|}
    \end{aligned}
\end{equation}
where $\mathrm{BiRetention}$ signifies bidirectional modeling.

\paragraph{From One-dimensional to Two-dimensional Decay:} While retention now supports bi-directional modeling, this capability remains confined to a one-dimensional level and is inadequate for two-dimensional images. To address this limitation, we extend the one-dimensional retention to encompass two dimensions.

In the context of images, each token is uniquely positioned with a two-dimensional coordinate within the plane, denoted as $(x_n, y_n)$ for the $n$-th token. To adapt to this, we adjust each element in the matrix $D$ to represent the Manhattan distance between the respective token pairs based on their 2D coordinates.
%\paragraph{One-dimensional Decay to Two-dimensional Decay.}  retention now has the ability for bi-directional modeling, this modeling capability remains limited to a one-dimensional level and is still not applicable to two-dimensional images. Therefore, we further extend the one-dimensional retention to two dimensions.
%For images, each token has a unique two-dimensional coordinate within the plane. For the $n$th token, we use $(x_n, y_n)$ to represent its two-dimensional coordinate. Based on the 2D coordinates of each token, we modify each element in the matrix $D$ to be the Manhattan distance between the corresponding token pairs. 
The matrix $D$ is redefined as follows:
\begin{equation}
    \label{eq:biret:2d}
    \begin{aligned}
        D_{nm}^{2d}=\gamma^{|x_n-x_m|+|y_n-y_m|}
    \end{aligned}
\end{equation}

\begin{figure}[ht]
    \centering
    \includegraphics[width=0.95\linewidth]{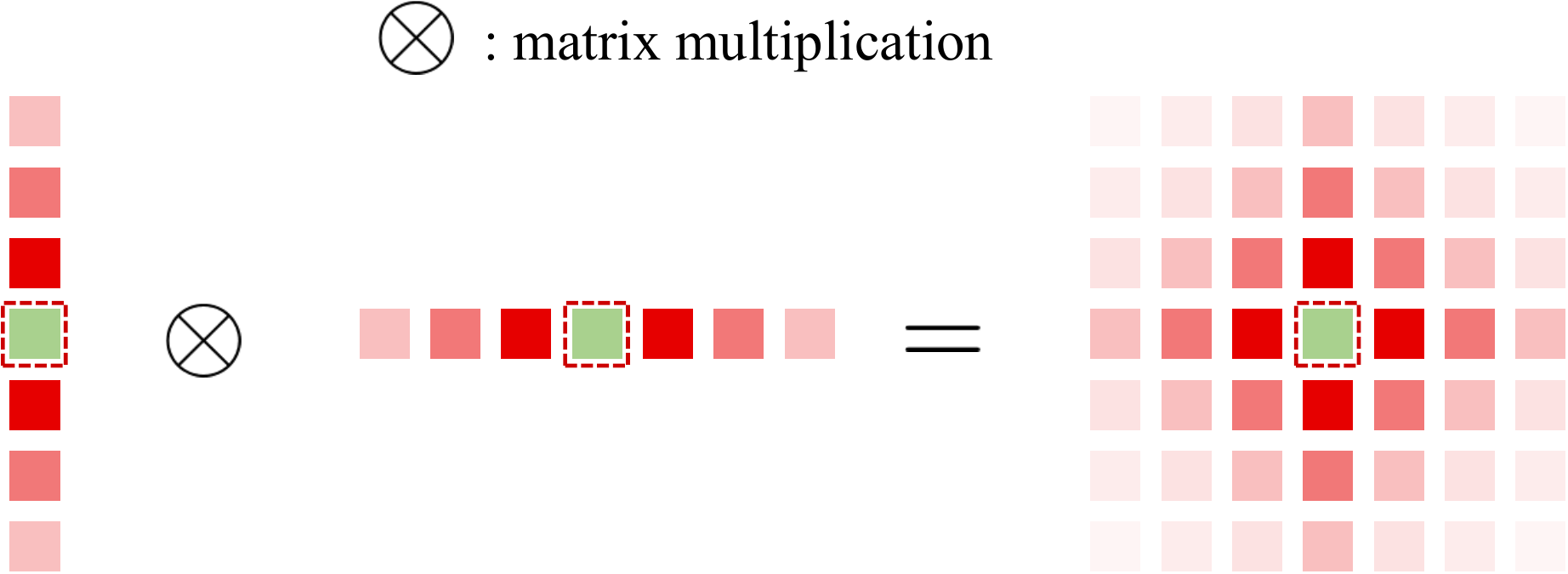}
    \caption{Spatial decay matrix in the decomposed MaSA.}
    \label{fig:decop}
\end{figure}

In the retention, the $\mathrm{Softmax}$ is abandoned and replaced with a gating function. This variation gives RetNet multiple flexible computation forms, enabling it to adapt to parallel training and recurrent inference processes. Despite this flexibility, when exclusively utilizing RetNet's parallel computation form in our experiments, the necessity of retaining the gating function becomes debatable. Our findings indicate that this modification does not improve results for vision models; instead, it introduces extra parameters and computational complexity. Consequently, we continue to employ $\mathrm{Softmax}$ to introduce nonlinearity to our model.
Combining the aforementioned steps, our Manhattan Self-Attention is expressed as 
\begin{equation}
    \label{eq:ReSA}
    \begin{aligned}
        \mathrm{MaSA}(X) &= (\mathrm{Softmax}(Q K^\intercal )\odot D^{2d})V \\
        D_{nm}^{2d}&=\gamma^{|x_n-x_m|+|y_n-y_m|}
    \end{aligned}
\end{equation}

\paragraph{Decomposed Manhattan Self-Attention.} In the early stages of the vision backbone, an abundance of tokens leads to high computational costs for Self-Attention when attempting to model global information. Our MaSA encounters this challenge as well. Utilizing existing sparse attention mechanisms~\cite{SwinTransformer, cswin, NAT, pvt, focal}, or the original RetNet's recurrent/chunk-wise recurrent form directly, disrupts the spatial decay matrix based on Manhattan distance, resulting in the loss of explicit spatial prior. To address this, we introduce a simple decomposition method that not only decomposes Self-Attention but also decomposes the spatial decay matrix.
The decomposed MaSA is represented in Eq.~\ref{eq:attnscore}. Specifically, we calculate attention scores separately for the horizontal and vertical directions in the image. Subsequently, we apply the one-dimensional bidirectional decay matrix to these attention weights. The one-dimensional decay matrix signifies the horizontal and vertical distances between tokens ($D^H_{nm}=\gamma^{|y_n-y_m|}$, $D^W_{nm}=\gamma^{|x_n-x_m|}$):
\begin{equation}
    \label{eq:attnscore}
    \begin{aligned}
        %&Q_H, K_H=(Q, K)^{\mathrm{B,L,C \xrightarrow{}B,W,H,C}}\\
        %&Q_W, K_W=(Q, K)^{\mathrm{B,L,C \xrightarrow{}B,H,W,C}}\\
        &Attn_H=\mathrm{Softmax}(Q_H K_H^\intercal) \odot D^{H},\\
        &Attn_W=\mathrm{Softmax}(Q_W K_W^\intercal) \odot D^{W}, \\
        &\mathrm{MaSA}(X)=Attn_H(Attn_WV)^\intercal \\
    \end{aligned}
\end{equation}

Based on the decomposition of MaSA, the shape of the receptive field of each token is shown in Fig.~\ref{fig:decop}, which is identical to the shape of the complete MaSA's receptive field. Fig.~\ref{fig:decop} indicates that our decomposition method fully preserves the explicit spatial prior.

\begin{table*}[ht]
    \centering
    \setlength{\tabcolsep}{2.8mm}
    \subfloat{
    \scalebox{0.85}{
    \begin{tabular}{c|c|c c|c}
        \toprule[1pt]
        \makecell{Cost} & Model & \makecell{Parmas\\(M)} & \makecell{FLOPs\\(G)} & \makecell{Top1-acc\\(\%)}\\
        \midrule[0.5pt]
        \multirow{14}{*}{\rotatebox{90}{\makecell{tiny model\\$\sim 2.5$G}}} 
        &PVTv2-b1~\cite{pvtv2} & 13 & 2.1 & 78.7 \\
        &QuadTree-B-b1~\cite{quadtree} & 14 & 2.3 & 80.0 \\
        &RegionViT-T~\cite{regionvit} & 14 & 2.4 & 80.4 \\
        &MPViT-XS~\cite{mpvit} & 11 & 2.9 & 80.9 \\
        &tiny-MOAT-2~\cite{MOAT} & 10 & 2.3 & 81.0 \\
        &VAN-B1~\cite{VAN} & 14 & 2.5 & 81.1 \\
        &BiFormer-T~\cite{biformer} & 13 & 2.2 & 81.4 \\
        &Conv2Former-N~\cite{conv2former} & 15 & 2.2 & 81.5 \\
        &CrossFormer-T~\cite{crossformer} & 28 & 2.9 & 81.5 \\
        &NAT-M~\cite{NAT} & 20 & 2.7 & 81.8 \\
        &QnA-T~\cite{QnA} & 16 & 2.5 & 82.0 \\
        &GC-ViT-XT~\cite{globalvit} & 20 & 2.6 & 82.0 \\
        &SMT-T~\cite{SMT} & 12 & 2.4 & 82.2 \\
        &\cellcolor{gray!30}RMT-T & \cellcolor{gray!30}14 & \cellcolor{gray!30}2.5 & \cellcolor{gray!30}82.4 \\
        \midrule[0.5pt]
        \multirow{25}{*}{\rotatebox{90}{\makecell{small model\\$\sim 4.5$G}}} 
        &DeiT-S~\cite{deit} & 22 & 4.6 & 79.9 \\
        &Swin-T~\cite{SwinTransformer} & 29 & 4.5 & 81.3 \\
        &ConvNeXt-T~\cite{convnext} & 29 & 4.5 & 82.1 \\
        &Focal-T~\cite{focal} & 29 & 4.9 & 82.2 \\
        %&InceptionNeXt-T~\cite{inceptionnext} & 28 & 4.2 & 82.3 \\
        &FocalNet-T~\cite{focalnet} & 29 & 4.5 & 82.3 \\
        &RegionViT-S~\cite{regionvit} & 31 & 5.3 & 82.6 \\
        &CSWin-T~\cite{cswin} & 23 & 4.3 & 82.7 \\
        &MPViT-S~\cite{mpvit} & 23 & 4.7 & 83.0 \\
        &ScalableViT-S~\cite{ScalableViT} & 32 & 4.2 & 83.1\\
        &SG-Former-S~\cite{sgformer} & 23 & 4.8 & 83.2 \\
        &MOAT-0~\cite{MOAT} & 28 & 5.7 & 83.3 \\
        &Ortho-S~\cite{Ortho} & 24 & 4.5 & 83.4 \\
        &InternImage-T~\cite{internimage} & 30 & 5.0 & 83.5 \\
        %&GC-ViT-T~\cite{globalvit} & 28 & 4.7 & 83.5 \\
        &CMT-S~\cite{cmt} & 25 & 4.0 & 83.5 \\
        %&SViT-S~\cite{stvit} & 25 & 4.4 & 83.6 \\
        &MaxViT-T~\cite{maxvit} & 31 & 5.6 & 83.6 \\
        %&FAT-B3~\cite{FAT} & 29 & 4.4 & 83.6 \\
        &SMT-S~\cite{SMT} & 20 & 4.8 & 83.7 \\
        &BiFormer-S~\cite{biformer} & 26 & 4.5 & 83.8 \\
        &\cellcolor{gray!30}RMT-S & \cellcolor{gray!30}27 & \cellcolor{gray!30}4.5 & \cellcolor{gray!30}84.1 \\
        \cline{2-5}
         &LV-ViT-S*~\cite{tokenlabel} & 26 & 6.6 & 83.3 \\
         &UniFormer-S*~\cite{uniformer} & 24 & 4.2 & 83.4 \\
         &WaveViT-S*~\cite{wavevit} & 23 & 4.7 & 83.9 \\
         &Dual-ViT-S*~\cite{dualvit2022} & 25 & 5.4 & 84.1 \\
         &VOLO-D1*~\cite{volo} & 27 & 6.8 & 84.2 \\
         &BiFormer-S*~\cite{biformer} & 26 & 4.5 & 84.3 \\
         &\cellcolor{gray!30}RMT-S* & \cellcolor{gray!30}27 & \cellcolor{gray!30}4.5 & \cellcolor{gray!30}84.8 \\
        \bottomrule[1pt]
    \end{tabular}}}
    \setlength{\tabcolsep}{2.8mm}
    \subfloat{
    \scalebox{0.85}{
    \begin{tabular}{c|c|c c|c}
        \toprule[1pt]
        \makecell{Cost} & Model & \makecell{Parmas\\(M)} & \makecell{FLOPs\\(G)} & \makecell{Top1-acc\\(\%)}\\
        \midrule[0.5pt]
        \multirow{20}{*}{\rotatebox{90}{\makecell{base model\\$\sim 9.0$G}}}
        &Swin-S~\cite{SwinTransformer} & 50 & 8.7 & 83.0 \\
        &ConvNeXt-S~\cite{convnext} & 50 & 8.7 & 83.1 \\
        &CrossFormer-B~\cite{crossformer} & 52 & 9.2 & 83.4 \\
        %&InceptionNeXt-S~\cite{inceptionnext} & 49 & 8.4 & 83.5 \\
        %&PVTv2-b4~\cite{pvtv2} & 63 & 10.0 & 83.6 \\
        &NAT-S~\cite{NAT} & 51 & 7.8 & 83.7 \\
        &Quadtree-B-b4~\cite{quadtree} & 64 & 11.5 & 84.0 \\
        &Ortho-B~\cite{Ortho} & 50 & 8.6 & 84.0 \\
        &ScaleViT-B~\cite{ScalableViT} & 81 & 8.6 & 84.1 \\
        &MOAT-1~\cite{MOAT} & 42 & 9.1 & 84.2 \\
        &InternImage-S~\cite{internimage} & 50 & 8.0 & 84.2 \\
        &DaViT-S~\cite{davit} & 50 & 8.8 & 84.2 \\
        &GC-ViT-S~\cite{globalvit} & 51 & 8.5 & 84.3 \\
        &BiFormer-B~\cite{biformer} & 57 & 9.8 & 84.3 \\
        &MViTv2-B~\cite{mvitv2} & 52 & 10.2 & 84.4 \\
        %&CMT-B~\cite{cmt} & 46 & 9.3 & 84.5 \\
        &iFormer-B~\cite{iformer} & 48 & 9.4 & 84.6 \\
        % &SViT-B~\cite{stvit} & 52 & 9.9 & 84.8 \\
        &\cellcolor{gray!30}RMT-B & \cellcolor{gray!30}54 & \cellcolor{gray!30}9.7 & \cellcolor{gray!30}85.0 \\
        \cline{2-5}
        %& \textcolor{gray!70}{LV-ViT-M*}~\cite{tokenlabel} & \textcolor{gray!70}{56} & \textcolor{gray!70}{16.0} & \textcolor{gray!70}{84.1} \\
         &WaveViT-B*~\cite{wavevit} & 34 & 7.2 & 84.8 \\
         &UniFormer-B*~\cite{uniformer} & 50 & 8.3 & 85.1 \\
         %&\textcolor{gray!70}{VOLO-D2*}~\cite{volo} & \textcolor{gray!70}{59} & \textcolor{gray!70}{14.1} & \textcolor{gray!70}{85.2} \\
         &Dual-ViT-B*~\cite{dualvit2022} & 43 & 9.3 & 85.2 \\
         &BiFormer-B*~\cite{biformer} & 58 & 9.8 & 85.4 \\
         &\cellcolor{gray!30}RMT-B* & \cellcolor{gray!30}55 & \cellcolor{gray!30}9.7 & \cellcolor{gray!30}85.6\\
        \midrule[0.5pt]
        \multirow{19}{*}{\rotatebox{90}{\makecell{large model\\$\sim 18.0$G}}}
        %&DeiT-B~\cite{deit} & 86 & 17.5 & 81.8 \\
        &Swin-B~\cite{SwinTransformer} & 88 & 15.4 & 83.3 \\
        &CaiT-M24~\cite{CaiT} & 186 & 36 & 83.4 \\
        &LITv2~\cite{litv2} & 87 & 13.2 & 83.6 \\
        &CrossFormer-L~\cite{crossformer} & 92 & 16.1 & 84.0 \\
        &Ortho-L~\cite{Ortho} & 88 & 15.4 & 84.2 \\
        &CSwin-B~\cite{cswin} & 78 & 15.0 & 84.2 \\
        %&MPViT-B~\cite{mpvit} & 75 & 16.4 & 84.3 \\
        %&ScalableViT-L~\cite{ScalableViT} & 104 & 14.7 & 84.4 \\
        &SMT-L~\cite{SMT} & 81 & 17.7 & 84.6 \\
        %&DaViT-B~\cite{davit} & 88 & 15.5 & 84.6 \\
        &MOAT-2~\cite{MOAT} & 73 & 17.2 & 84.7 \\
        &SG-Former-B~\cite{sgformer} & 78 & 15.6 & 84.7 \\
        &iFormer-L~\cite{iformer} & 87 & 14.0 & 84.8 \\
        %&CMT-L~\cite{cmt} & 75 & 19.5 & 84.8 \\
        &InterImage-B~\cite{internimage} & 97 & 16.0 & 84.9 \\
        &MaxViT-B~\cite{maxvit} & 120 & 23.4 & 84.9 \\
        &GC-ViT-B~\cite{globalvit} & 90 & 14.8 & 85.0 \\
        % &SViT-L~\cite{stvit} & 95 & 15.6 & 85.3 \\
        &\cellcolor{gray!30}RMT-L & \cellcolor{gray!30}95 & \cellcolor{gray!30}18.2 & \cellcolor{gray!30}85.5 \\
        \cline{2-5}
        %& \textcolor{gray!70}{LV-ViT-L*}~\cite{tokenlabel} & \textcolor{gray!70}{150} & \textcolor{gray!70}{59.0} & \textcolor{gray!70}{85.3} \\
         & VOLO-D3*~\cite{volo} & 86 & 20.6 & 85.4 \\
         & WaveViT-L*~\cite{wavevit} & 58 & 14.8 & 85.5 \\
         & UniFormer-L*~\cite{uniformer} & 100 & 12.6 & 85.6 \\
         & Dual-ViT-L*~\cite{dualvit2022} & 73 & 18.0 & 85.7 \\
         &\cellcolor{gray!30}RMT-L* & \cellcolor{gray!30}96 & \cellcolor{gray!30}18.2 & \cellcolor{gray!30}86.1\\
        \bottomrule[1pt]
    \end{tabular}}}
    \caption{Comparison with the state-of-the-art on ImageNet-1K classification. ``*" indicates the model trained with token labeling~\cite{tokenlabel}.}
    \vspace{-3mm}
    \label{tab:ImageNet}
\end{table*}

To further enhance the local expression capability of MaSA, following  \cite{biformer}, we introduce a Local Context Enhancement module using DWConv:
\begin{equation}
    \label{lce}
    \begin{aligned}
        X_{out}=\mathrm{MaSA}(X)+\mathrm{LCE}(V);
    \end{aligned}
\end{equation}

\subsection{Overall Architecture}

We construct the RMT based on MaSA, and its architecture is illustrated in Fig.~\ref{fig:pipeline}. Similar to previous general vision backbones~\cite{pvt, pvtv2, SwinTransformer, ViL}, RMT is divided into four stages. The first three stages utilize the decomposed MaSA, while the last uses the original MaSA. Like many previous backbones~\cite{cmt, biformer, uniformer, vsa}, we incorporate CPE~\cite{CPVT} into our model. 
\begin{table*}[!ht]
    \setlength{\tabcolsep}{1.25mm}
    \centering
    \scalebox{0.85}{
    \begin{tabular}{c|c c|c c c c c c|c c|c c c c c c}
        \toprule[1pt]
        \multirow{2}{*}{Backbone} & \multirow{2}{*}{\makecell{Params\\(M)}} & \multirow{2}{*}{\makecell{FLOPs\\(G)}} & \multicolumn{6}{c|}{Mask R-CNN $1\times$} & \multirow{2}{*}{\makecell{Params\\(M)}} & \multirow{2}{*}{\makecell{FLOPs\\(G)}} & \multicolumn{6}{c}{RetinaNet $1\times$}\\
         & & & $AP^b$ & $AP^b_{50}$ & $AP^b_{75}$ & $AP^m$ & $AP^m_{50}$ & $AP^m_{75}$ & & & $AP^b$ & $AP^b_{50}$ & $AP^b_{75}$ & $AP^b_S$ & $AP^b_{M}$ & $AP^b_{L}$ \\
         \midrule[0.5pt]
        PVT-T~\cite{pvt} & 33 & 240 & 39.8 & 62.2 & 43.0 & 37.4 & 59.3 & 39.9 & 23 & 221 & 39.4 & 59.8 & 42.0 & 25.5 & 42.0 & 52.1 \\
        PVTv2-B1~\cite{pvtv2} & 33 & 243 & 41.8 & 54.3 & 45.9 & 38.8 & 61.2 & 41.6 & 23 & 225 & 41.2 & 61.9 & 43.9 & 25.4 & 44.5 & 54.3 \\
        MPViT-XS~\cite{mpvit} & 30 & 231 & 44.2 & 66.7 & 48.4 & 40.4 & 63.4 & 43.4 & 20 & 211 & 43.8 & 65.0 & 47.1 & 28.1 & 47.6 & 56.5 \\
        %FAT-B2~\cite{FAT} & 33 & 215 & 45.2 & 67.9 & 49.0 & 41.3 & 64.6 & 44.0 & 23 & 196 & 44.0 & 65.2 & 47.2 & 27.5 & 47.7 & 58.8 \\
        \rowcolor{gray!30}RMT-T & 33 & 218 & \textbf{47.1} & \textbf{68.8} & \textbf{51.7} & \textbf{42.6} & \textbf{65.8} & \textbf{45.9} & 23 & 199 & \textbf{45.1} & \textbf{66.2} & \textbf{48.1} & \textbf{28.8} & \textbf{48.9} & \textbf{61.1} \\
        \midrule[0.5pt]
        %ResNet-50~\cite{resnet} & 44 & 260 & 38.0 & 58.6 & 41.4 & 34.4 & 55.1 & 36.7 & 38 & 239 & 36.3 & 55.3 & 38.6 & 19.3 & 40.4 & 48.8 \\
        Swin-T~\cite{SwinTransformer} & 48 & 267 & 43.7 & 66.6 & 47.7 & 39.8 & 63.3 & 42.7 & 38 & 248 & 41.7 & 63.1 & 44.3 & 27.0 & 45.3 & 54.7 \\
        CMT-S~\cite{cmt} & 45 & 249 & 44.6 & 66.8 & 48.9 & 40.7 & 63.9 & 43.4 & 44 & 231 & 44.3 & 65.5 & 47.5 & 27.1 & 48.3 & 59.1 \\
        CrossFormer-S~\cite{crossformer} & 50 & 301 & 45.4 & 68.0 & 49.7 & 41.4 & 64.8 & 44.6 & 41 & 272 & 44.4 & 65.8 & 47.4 & 28.2 & 48.4 & 59.4 \\
        ScalableViT-S~\cite{ScalableViT} & 46 & 256 & 45.8 & 67.6 & 50.0 & 41.7 & 64.7 & 44.8 & 36 & 238 & 45.2 & 66.5 & 48.4 & 29.2 & 49.1 & 60.3 \\
        MPViT-S~\cite{mpvit} & 43 & 268 & 46.4 & 68.6 & 51.2 & 42.4 & 65.6 & 45.7 & 32 & 248 & 45.7 & 57.3 & 48.8 & 28.7 & 49.7 & 59.2 \\
        %Dual-ViT-S*~\cite{dualvit2022} & -- & -- & 46.5 & 68.3 & 51.2 & 42.2 & 65.3 & 46.1 & -- & -- & 46.2 & 67.4 & 49.9 & 30.6 & 49.9 & 60.9 \\
        CSWin-T~\cite{cswin} & 42 & 279 & 46.7 & 68.6 & 51.3 & 42.2 & 65.6 & 45.4 & -- & -- & -- & -- & -- & -- & -- & -- \\
        InternImage-T~\cite{internimage} & 49 & 270 & 47.2 & 69.0 & 52.1  & 42.5 & 66.1 & 45.8 & -- & -- & -- & -- & -- & -- & -- & -- \\
        %SViT-S~\cite{stvit} & 44 & 252 & 47.6 & 70.0 & 52.3 & 43.1 & 66.8 & 46.5 & -- & -- & -- & -- & -- & -- & -- & -- \\
        %FAT-B3~\cite{FAT} & 49 & -- & 47.6 & 69.7 & 52.3 & 43.1 & 66.4 & 46.2 & 39 & -- & 45.9 & 66.9 & 49.5 & 29.3 & 50.1 & 60.9 \\
        SMT-S~\cite{SMT} & 40 & 265 & 47.8 & 69.5 & 52.1 & 43.0 & 66.6 & 46.1 & -- & -- & -- & -- & -- & -- & -- & -- \\
        BiFormer-S~\cite{biformer} & -- & -- & 47.8 & 69.8 & 52.3 & 43.2 & 66.8 & 46.5 & -- & -- & 45.9 & 66.9 & 49.4 & 30.2 & 49.6 & 61.7 \\
        \rowcolor{gray!30}RMT-S & 46 & 262 & \textbf{49.0} & \textbf{70.8} & \textbf{53.9} & \textbf{43.9} & \textbf{67.8} & \textbf{47.4} & 36 & 244 & \textbf{47.8} & \textbf{69.1} & \textbf{51.8} & \textbf{32.1} & \textbf{51.8} & \textbf{63.5} \\
        \midrule[0.5pt]
        ResNet-101~\cite{resnet} & 63 & 336 & 40.4 & 61.1 & 44.2 & 36.4 & 57.7 & 38.8 & 58 & 315 & 38.5 & 57.8 & 41.2 & 21.4 & 42.6 & 51.1 \\
        Swin-S~\cite{SwinTransformer} & 69 & 359 & 45.7 & 67.9 & 50.4 & 41.1 & 64.9 & 44.2 & 60 & 339 & 44.5 & 66.1 & 47.4 & 29.8 & 48.5 & 59.1 \\
        ScalableViT-B~\cite{ScalableViT} &95 & 349 & 46.8 & 68.7 & 51.5 & 42.5 & 65.8 & 45.9 & 85 & 330 & 45.8 & 67.3 & 49.2 & 29.9 & 49.5 & 61.0 \\
        InternImage-S~\cite{internimage} & 69 & 340 & 47.8 & 69.8 & 52.8 & 43.3 & 67.1 & 46.7 & -- & -- & -- & -- & -- & -- & -- & -- \\
        CSWin-S~\cite{cswin} & 54 & 342 & 47.9 & 70.1 & 52.6 & 43.2 & 67.1 & 46.2 & -- & -- & -- & -- & -- & -- & -- & -- \\
        %Dual-ViT-B*~\cite{dualvit2022} & -- & -- & 48.4 & 69.9 & 53.3 & 43.4 & 66.7 & 46.8 & -- & -- & 47.4 & 68.1 & 51.2 & 29.6 & 51.9 & 63.1 \\
        %SViT-B~\cite{stvit} & 70 & 359 & 49.7 & 71.7 & 54.7 & 44.8 & 68.9 & 48.7 & -- & -- & -- & -- & -- & -- & -- & -- \\
        BiFormer-B~\cite{biformer} & -- & -- & 48.6 & 70.5 & 53.8 & 43.7 & 67.6 & 47.1 & -- & -- & 47.1 & 68.5 & 50.4 & 31.3 & 50.8 & 62.6 \\
        \rowcolor{gray!30}RMT-B & 73 & 373 & \textbf{51.1} & \textbf{72.5} & \textbf{56.1} & \textbf{45.5} & \textbf{69.7} & \textbf{49.3} & 63 & 355 & \textbf{49.1} & \textbf{70.3} & \textbf{53.0} & \textbf{32.9} & \textbf{53.2} & \textbf{64.2} \\
        \midrule[0.5pt]
        Swin-B~\cite{SwinTransformer} & 107 & 496 & 46.9 & 69.2 & 51.6 & 42.3 & 66.0 & 45.5 & 98 & 477 & 45.0 & 66.4 & 48.3 & 28.4 & 49.1 & 60.6 \\
        PVTv2-B5~\cite{pvtv2} & 102 & 557 & 47.4 & 68.6 & 51.9 & 42.5 & 65.7 & 46.0 & -- & -- & -- & -- & -- & -- & -- & -- \\
        Focal-B~\cite{focal} & 110 & 533 & 47.8 & 70.2 & 52.5 & 43.2 & 67.3 & 46.5 & 101 & 514 & 46.3 & 68.0 & 49.8 & 31.7 & 50.4 & 60.8 \\
        MPViT-B~\cite{mpvit} & 95 & 503 & 48.2 & 70.0 & 52.9 & 43.5 & 67.1 & 46.8 & 85 & 482 & 47.0 & 68.4 & 50.8 & 29.4 & 51.3 & 61.5 \\
        CSwin-B~\cite{cswin} & 97 & 526 & 48.7 & 70.4 & 53.9 & 43.9 & 67.8 & 47.3 & -- & -- & -- & -- & -- & -- & -- & -- \\
        InternImage-B~\cite{internimage} & 115 & 501 & 48.8 & 70.9 & 54.0 & 44.0 & 67.8 & 47.4 & -- & -- & -- & -- & -- & -- & -- & -- \\
        \rowcolor{gray!30}RMT-L & 114 & 557 & \textbf{51.6} & \textbf{73.1} & \textbf{56.5} & \textbf{45.9} & \textbf{70.3} & \textbf{49.8} & 104 & 537 & \textbf{49.4} & \textbf{70.6} & \textbf{53.1} & \textbf{34.2} & \textbf{53.9} & \textbf{65.2} \\
        \bottomrule[1pt]
    \end{tabular}}
    \caption{Comparison to other backbones using RetinaNet and Mask R-CNN on COCO val2017 object detection and instance segmentation. }
    \vspace{-3mm}
    \label{tab:COCO1x}
\end{table*}

\section{Experiments}

We conducted extensive experiments on multiple vision tasks, such as image classification on ImageNet-1K~\cite{imagenet}, object detection and instance segmentation on COCO 2017~\cite{coco}, and semantic segmentation on ADE20K~\cite{ade20k}. We also make ablation studies to validate the importance of each component in RMT. More details can be found in \textcolor{red}{Appendix}.

\subsection{Image Classification}

\paragraph{Settings.}We train our models on ImageNet-1K~\cite{imagenet} from scratch. We follow the same training strategy in \cite{deit}, with the only supervision being classification loss for a fair comparison. The maximum rates of increasing stochastic depth~\cite{droppath} are set to 0.1/0.15/0.4/0.5 for RMT-T/S/B/L~\cite{droppath}, respectively. We use the AdamW optimizer with a cosine decay learning rate scheduler to train the models. We set the initial learning rate, weight decay, and batch size to 0.001, 0.05, and 1024, respectively. We adopt the strong data augmentation and regularization used in \cite{SwinTransformer}. Our settings are RandAugment~\cite{randomaugment} (randm9-mstd0.5-inc1), Mixup~\cite{mixup} (prob=0.8), CutMix~\cite{cutmix} (prob=1.0), Random Erasing~\cite{randera} (prob=0.25). In addition to the conventional training methods, similar to LV-ViT~\cite{tokenlabel} and VOLO~\cite{volo}, we train a model that utilizes token labeling to provide supplementary supervision.

\paragraph{Results.}We compare RMT against many state-of-the-art models in Tab.~\ref{tab:ImageNet}. Results in the table demonstrate that RMT consistently outperforms previous models across all settings. Specifically, RMT-S achieves \textbf{84.1\%} Top1-accuracy with only \textbf{4.5} GFLOPs. RMT-B also surpasses iFormer~\cite{iformer} by \textbf{0.4\%} with similar FLOPs. Furthermore, our RMT-L model surpasses MaxViT-B~\cite{maxvit} in top1-accuracy by \textbf{0.6\%} while using fewer FLOPs. Our RMT-T has also outperformed many lightweight models. As for the model trained using token labeling, our RMT-S outperforms the current state-of-the-art BiFormer-S by \textbf{0.5\%}.

\subsection{Object Detection and Instance Segmentation}

\begin{table}[h]
    \centering
    \setlength{\tabcolsep}{0.22mm}
    \scalebox{0.82}{
    \begin{tabular}{c|c c|c c c c c c}
        \toprule[1pt]
         \multirow{2}{*}{Backbone} & \multirow{2}{*}{\makecell{Params\\(M)}} & \multirow{2}{*}{\makecell{FLOPs\\(G)}} & \multicolumn{6}{c}{Mask R-CNN $3\times$+MS}\\
          & & & $AP^b$ & $AP^b_{50}$ & $AP^b_{75}$ & $AP^m$ & $AP^m_{50}$ & $AP^m_{75}$\\
          \midrule[0.5pt]
          %Swin-T~\cite{SwinTransformer} & 48 & 267 & 46.0 & 68.1 & 50.3 & 41.6 & 65.1 & 44.9 \\
          ConvNeXt-T~\cite{convnext} & 48 & 262 & 46.2 & 67.9 & 50.8 & 41.7 & 65.0 & 45.0 \\
          Focal-T~\cite{focal} & 49 & 291 & 47.2 & 69.4 & 51.9 & 42.7 & 66.5 & 45.9 \\
          NAT-T~\cite{NAT} & 48 & 258 & 47.8 & 69.0 & 52.6 & 42.6 & 66.0 & 45.9 \\
          GC-ViT-T~\cite{globalvit} & 48 & 291 & 47.9 & 70.1 & 52.8 & 43.2 & 67.0 & 46.7 \\
          MPViT-S~\cite{mpvit} & 43 & 268 & 48.4 & 70.5 & 52.6 & 43.9 & 67.6 & 47.5 \\
          Ortho-S~\cite{Ortho} & 44 & 277 & 48.7 & 70.5 & 53.3 & 43.6 & 67.3 & 47.3 \\
          SMT-S~\cite{SMT} & 40 & 265 & 49.0 & 70.1 & 53.4 & 43.4 & 67.3 & 46.7\\
          CSWin-T~\cite{cswin} & 42 & 279 & 49.0 & 70.7 & 53.7 & 43.6 & 67.9 & 46.6\\
          InternImage-T~\cite{internimage} & 49 & 270 & 49.1 & 70.4 & 54.1 & 43.7 & 67.3 & 47.3 \\
          \rowcolor{gray!30}RMT-S & 46 & 262 & \textbf{50.7} & \textbf{71.9} & \textbf{55.6} & \textbf{44.9} & \textbf{69.1} & \textbf{48.4}\\
          \midrule[0.5pt]
          ConvNeXt-S~\cite{convnext} & 70 & 348 & 47.9 & 70.0 & 52.7 & 42.9 & 66.9 & 46.2 \\
          NAT-S~\cite{NAT} & 70 & 330 & 48.4 & 69.8 & 53.2 & 43.2 & 66.9 & 46.4 \\
          Swin-S~\cite{SwinTransformer} & 69 & 359 & 48.5 & 70.2 & 53.5 & 43.3 & 67.3 & 46.6 \\
          InternImage-S~\cite{internimage} & 69 & 340 & 49.7 & 71.1 & 54.5 & 44.5 & 68.5 & 47.8 \\
          SMT-B~\cite{SMT} & 52 & 328 & 49.8 & 71.0 & 54.4 & 44.0 & 68.0 & 47.3\\
          CSWin-S~\cite{cswin} & 54 & 342 & 50.0 & 71.3 & 54.7 & 44.5 & 68.4 & 47.7 \\
          \rowcolor{gray!30}RMT-B & 73 & 373 & \textbf{52.2} & \textbf{72.9} & \textbf{57.0} & \textbf{46.1} & \textbf{70.4} & \textbf{49.9}  \\
          \bottomrule
    \end{tabular}}
    \caption{Comparison to other backbones using Mask R-CNN with "$3\times+\mathrm{MS}$" schedule.}
    \vspace{-5mm}
    \label{tab:COCO3x}
\end{table}

\begin{table}[h]
    \centering
    \setlength{\tabcolsep}{0.22mm}
    \scalebox{0.82}{
    \begin{tabular}{c|c c|c c c c c c}
        \toprule[1pt]
         \multirow{2}{*}{Backbone} & \multirow{2}{*}{\makecell{Params\\(M)}} & \multirow{2}{*}{\makecell{FLOPs\\(G)}} & \multicolumn{6}{c}{Cascade Mask R-CNN $3\times$+MS}\\
          & & & $AP^b$ & $AP^b_{50}$ & $AP^b_{75}$ & $AP^m$ & $AP^m_{50}$ & $AP^m_{75}$\\
          \midrule[0.5pt]
          Swin-T~\cite{SwinTransformer} & 86 & 745 & 50.5 & 69.3 & 54.9 & 43.7 & 66.6 & 47.1 \\
          NAT-T~\cite{NAT} & 85 & 737 & 51.4 & 70.0 & 55.9 & 44.5 & 67.6 & 47.9 \\
          GC-ViT-T~\cite{globalvit} & 85 & 770 & 51.6 & 70.4 & 56.1 & 44.6 & 67.8 & 48.3 \\
          SMT-S~\cite{SMT} & 78 & 744 & 51.9 & 70.5 & 56.3 & 44.7 & 67.8 & 48.6 \\
          UniFormer-S~\cite{uniformer} & 79 & 747 & 52.1 & 71.1 & 56.6 & 45.2 & 68.3 & 48.9 \\
          Ortho-S~\cite{Ortho} & 81 & 755 & 52.3 & 71.3 & 56.8 & 45.3 & 68.6 & 49.2 \\
          HorNet-T~\cite{hornet} & 80 & 728 & 52.4 & 71.6 & 56.8 & 45.6 & 69.1 & 49.6 \\
          CSWin-T~\cite{cswin} & 80 & 757 & 52.5 & 71.5 & 57.1 & 45.3 & 68.8 & 48.9 \\
          \rowcolor{gray!30}RMT-S & 83 & 741 & \textbf{53.2} & \textbf{72.0} & \textbf{57.8} & \textbf{46.1} & \textbf{69.8} & \textbf{49.8}\\
          \midrule[0.5pt]
          Swin-S~\cite{SwinTransformer} & 107 & 838 & 51.9 & 70.7 & 56.3 & 45.0 & 68.2 & 48.8 \\
          NAT-S~\cite{NAT} & 108 & 809 & 51.9 & 70.4 & 56.2 & 44.9 & 68.2 & 48.6 \\
          GC-ViT-S~\cite{globalvit} & 108 & 866 & 52.4 & 71.0 & 57.1 & 45.4 & 68.5 & 49.3\\
          DAT-S~\cite{dat} & 107 & 857 & 52.7 & 71.7 & 57.2 & 45.5 & 69.1 & 49.3 \\
          HorNet-S~\cite{hornet} & 108 & 827 & 53.3 & 72.3 & 57.8 & 46.3 & 69.9 & 50.4 \\
          CSWin-S~\cite{cswin} & 92 & 820 & 53.7 & 72.2 & 58.4 & 46.4 & 69.6 & 50.6 \\
          UniFormer-B~\cite{uniformer} & 107 & 878 & 53.8 & 72.8 & 58.5 & 46.4 & 69.9 & 50.4 \\
          \rowcolor{gray!30}RMT-B & 111 & 852 & \textbf{54.5} & \textbf{72.8} & \textbf{59.0} & \textbf{47.2} & \textbf{70.5} & \textbf{51.4}  \\
          \bottomrule
    \end{tabular}}
    \caption{Comparison to other backbones using Cascade Mask R-CNN with "$3\times+\mathrm{MS}$" schedule.}
    \vspace{-5mm}
    \label{tab:CASCOCO3x}
\end{table}

\paragraph{Settings.}We adopt MMDetection~\cite{mmdetection} to implement RetinaNet~\cite{retinanet}, Mask-RCNN~\cite{maskrcnn} and Cascade Mask R-CNN~\cite{cai18cascadercnn}. We use the commonly used ``$1\times$" (12 training epochs) setting for the RetinaNet and Mask R-CNN. Besides, we use ``$3\times+\mathrm{MS}$" for Mask R-CNN and Cascade Mask R-CNN. Following \cite{SwinTransformer}, during training, images are resized to the shorter side of 800 pixels while the longer side is within 1333 pixels. We adopt the AdamW optimizer with a learning rate of 0.0001 and batch size of 16 to optimize the model. For the ``$1\times$" schedule, the learning rate declines with the decay rate of 0.1 at the epoch 8 and 11. While for the ``$3\times+\mathrm{MS}$" schedule, the learning rate declines with the decay rate of 0.1 at the epoch 27 and 33.

\paragraph{Results.}Tab.~\ref{tab:COCO1x}, Tab.~\ref{tab:COCO3x} and Tab.~\ref{tab:CASCOCO3x} show the results with different detection frameworks. The results demonstrate that our RMT performs best in all comparisons. For the RetinaNet framework, our RMT-T outperforms MPViT-XS by \textbf{+1.3} AP, while S/B/L also perform better than other methods. As for the Mask R-CNN with ``$1\times$" schedule, RMT-L outperforms the recent InternImage-B by \textbf{+2.8} box AP and \textbf{+1.9} mask AP. For ``$3\times+\mathrm{MS}$" schedule, RMT-S outperforms InternImage-T for \textbf{+1.6} box AP and \textbf{+1.2} mask AP. Besides, regarding the Cascade Mask R-CNN, our RMT still performs much better than other backbones. All the above results tell that RMT outperforms its counterparts by evident margins.

\subsection{Semantic Segmentation}

\begin{table}[t]
    \centering
    \scalebox{0.76}{
    \begin{tabular}{c|c|c|c|c}
         \toprule[1pt]
         Backbone & Method & Params(M) & FLOPs(G) & mIoU(\%)\\
         \midrule[0.5pt]
         ResNet18~\cite{resnet}& FPN & 15.5 & 32.2 & 32.9 \\
         PVTv2-B1~\cite{pvtv2}& FPN & 17.8 & 34.2 & 42.5 \\
         VAN-B1~\cite{VAN}& FPN & 18.1 & 34.9 & 42.9 \\
         EdgeViT-S~\cite{edgevit}& FPN & 16.9 & 32.1 & 45.9\\
         \rowcolor{gray!30}RMT-T& FPN & 17.0 & 33.7 & \textbf{46.4}\\
         \midrule[0.5pt]
         DAT-T~\cite{dat}& FPN & 32 & 198 & 42.6 \\
         RegionViT-S+~\cite{regionvit} & FPN & 35 & 236 & 45.3 \\
         CrossFormer-S~\cite{crossformer}& FPN & 34 & 221 & 46.0\\
         UniFormer-S~\cite{uniformer}& FPN & 25 & 247 & 46.6\\
         %CSWin-T~\cite{cswin}& FPN & 26 & 202 & 48.2 \\
         Shuted-S~\cite{shunted}& FPN & 26 & 183 & 48.2 \\
         \rowcolor{gray!30}RMT-S& FPN & 30 & 180 & \textbf{49.4}\\
         \midrule[0.5pt]
         DAT-S~\cite{dat}& FPN & 53 & 320 & 46.1\\
         RegionViT-B+~\cite{regionvit} & FPN & 77 & 459 & 47.5 \\
         UniFormer-B~\cite{uniformer}& FPN & 54 & 350 & 47.7 \\
         CrossFormer-B~\cite{crossformer}& FPN & 56 & 331 & 47.7 \\
         CSWin-S~\cite{cswin}& FPN & 39 & 271 & 49.2 \\
         \rowcolor{gray!30}RMT-B& FPN & 57 & 294 & \textbf{50.4} \\
         \midrule[0.5pt]
         DAT-B~\cite{dat}& FPN & 92 & 481 & 47.0 \\
         CrossFormer-L~\cite{crossformer}& FPN & 95 & 497 & 48.7 \\
         CSWin-B~\cite{cswin}& FPN & 81 & 464 & 49.9 \\
         \rowcolor{gray!30}RMT-L& FPN & 98 & 482 & \textbf{51.4} \\
         \midrule[1pt]
         DAT-T~\cite{dat}& UperNet & 60 & 957 & 45.5 \\
         NAT-T~\cite{NAT}& UperNet & 58 & 934 & 47.1 \\
         InternImage-T~\cite{internimage}& UperNet & 59 & 944 & 47.9 \\
         MPViT-S~\cite{mpvit}& UperNet & 52 & 943 & 48.3 \\
         %HorNet-T~\cite{hornet}& UperNet & 55 & 924 & 49.2 \\
         SMT-S~\cite{SMT}& UperNet & 50 & 935 & 49.2 \\
         \rowcolor{gray!30}RMT-S& UperNet & 56 & 937 & \textbf{49.8} \\
         \midrule[0.5pt]
         DAT-S~\cite{dat}& UperNet & 81 & 1079 & 48.3\\
         SMT-B~\cite{SMT}& UperNet & 62 & 1004 & 49.6\\
         HorNet-S~\cite{hornet}& UperNet & 85 & 1027 & 50.0 \\
         InterImage-S~\cite{internimage}& UperNet & 80 & 1017 & 50.2\\
         MPViT-B~\cite{mpvit}& UperNet & 105 & 1186 & 50.3 \\
         CSWin-S~\cite{cswin}& UperNet & 65 & 1027 & 50.4\\
         \rowcolor{gray!30}RMT-B& UperNet & 83 & 1051 & \textbf{52.0}\\
         \midrule
         Swin-B~\cite{SwinTransformer} & UperNet & 121 & 1188 & 48.1 \\
         GC ViT-B~\cite{globalvit} & UperNet & 125 & 1348 & 49.2\\
         DAT-B~\cite{dat} & UperNet & 121 & 1212 & 49.4 \\
         InternImage-B~\cite{internimage} & UperNet & 128 & 1185 & 50.8\\
         CSWin-B~\cite{cswin} & UperNet & 109 & 1222 & 51.1\\
         \rowcolor{gray!30}RMT-L&UperNet & 125 & 1241 & \textbf{52.8}\\
         \bottomrule[1pt]
    \end{tabular}}
    \caption{Comparison with the state-of-the-art on ADE20K. }
    \vspace{-3mm}
    \label{tab:semanticfpn}
\end{table}

\paragraph{Settings.}We adopt the Semantic FPN~\cite{semanticfpn} and UperNet~\cite{upernet} based on MMSegmentation~\cite{mmsegmentation}, apply RMTs which are pretrained on ImageNet-1K as backbone. We use the same setting of PVT~\cite{pvt} to train the Semantic FPN, and we train the model for 80k iterations. All models are trained with the input resolution of $512\times512$. When testing the model, we resize the shorter side of the image to 512 pixels. As for UperNet, we follow the default settings in Swin~\cite{SwinTransformer}. We take AdamW with a weight decay of 0.01 as the optimizer to train the models for 160K iterations. The learning rate is set to $6\times10^{-5}$ with 1500 iterations warmup.  

\begin{table*}[t]
    \centering
    \setlength{\tabcolsep}{3.8mm}
    \scalebox{0.94}{
    \begin{tabular}{c|c c|c|c c|c}
        \toprule[1pt]
         Model & Params(M) & FLOPs(G) & Top1-acc(\%) & $AP^b$ & $AP^m$ & mIoU(\%)\\
         \midrule
         DeiT-S~\cite{deit} & 22 & 4.6 & 79.8 & -- & -- & -- \\
         \rowcolor{gray!30}RMT-DeiT-S & 22 & 4.6 & 81.7(\textcolor{red}{+1.9}) & -- & -- & -- \\
         \midrule
         Swin-T~\cite{SwinTransformer} & 29 & 4.5 & 81.3 & 43.7 & 39.8 & 44.5 \\
         %DAT-T~\cite{dat} & 29 & 4.6 & 82.0 & 44.4 & 40.4 & -- \\
         %Focal-T~\cite{focal} & 29 & 4.9 & 82.2 & 44.8 & 41.0 & -- \\
         \rowcolor{gray!30}RMT-Swin-T & 29 & 4.7 & 83.6(\textcolor{red}{+2.3}) & 47.8(\textcolor{red}{+4.1}) & 43.1(\textcolor{red}{+3.3}) & 49.1(\textcolor{red}{+4.6}) \\
         \midrule[0.5pt]
         Swin-S~\cite{SwinTransformer} & 50 & 8.8 & 83.0 & 45.7 & 41.1 & 47.6 \\
         %DAT-S~\cite{dat} & 50 & 9.0 & 83.7 & 47.1 & 42.5 & --\\
         %Focal-S~\cite{focal} & 51 & 9.1 & 83.5 & 47.4 & 42.8 & --\\
         \rowcolor{gray!30}RMT-Swin-S & 50 & 9.1 & 84.5(\textcolor{red}{+1.5})& 49.5(\textcolor{red}{+3.8}) & 44.2(\textcolor{red}{+3.1}) & 51.0 (\textcolor{red}{+3.4})\\
         \midrule[0.5pt]
         RMT-T & 14.3 & 2.5 & 82.4 & 47.1 & 42.6 & 46.4\\
         MaSA$\xrightarrow{}$Attention & 14.3 & 2.5 & 81.6(\textcolor{red}{-0.8}) & 44.6(\textcolor{red}{-2.5}) & 40.7(\textcolor{red}{-1.9}) & 43.9(\textcolor{red}{-2.5})\\
         Softmax$\xrightarrow{}$Gate & 15.6 & 2.7 & Nan & -- & -- & -- \\
         w/o LCE & 14.2 & 2.4 & 82.1 & 46.7 & 42.3 & 46.0 \\
         w/o CPE & 14.3 & 2.5 & 82.2 & 47.0 & 42.4 & 46.4 \\
         w/o Stem & 14.3 & 2.2 & 82.2 & 46.8 & 42.3 & 46.2 \\
         \bottomrule[1pt]
    \end{tabular}}
    \caption{Ablation study. We make a strict comparison among RMT, DeiT, and Swin-Transformer.}
    \vspace{-3mm}
    \label{tab:ablation_study}
\end{table*}

\begin{table}[t]
    \centering
    \setlength{\tabcolsep}{3.1mm}
    \scalebox{0.82}{
    \begin{tabular}{c|c c|c c}
    \toprule[1pt]
        3rd stage & FLOPs(G) & Top1(\%) & FLOPs(G) & mIoU(\%) \\
        \midrule[0.5pt]
        MaSA-d & 4.5 & 84.1 & 180 & 49.4 \\
        MaSA & 4.8 & 84.1 & 246 & 49.7 \\
    \bottomrule[1pt]
    \end{tabular}}
    \caption{Comparison between decomposed MaSA (MaSA-d) and original MaSA.}
    \vspace{-3mm}
    \label{tab:ab_decop}
\end{table}

\begin{table}[t]
    \centering
    \setlength{\tabcolsep}{2.9mm}
    \scalebox{0.86}{
    \begin{tabular}{c|c c c|c}
        \toprule[1pt]
        Method & \makecell{Params\\(M)} & \makecell{FLOPs$\downarrow$\\(G)} & \makecell{Throughput$\uparrow$\\(imgs/s)} & \makecell{Top1\\(\%)}\\
        \midrule
        Parallel & 27 & {10.9} & {262} & -- \\
        Chunklen\_4 & 27 & 4.5 & {192} & --\\
        Chunklen\_49 & 27 & 4.7 & {446} & 82.1\\
        Recurrent & 27 & 4.5 & {61}& --\\
        \midrule
        MaSA & 27 & 4.5 & {876} & {84.1} \\
        \bottomrule[1pt]
    \end{tabular}}
    \caption{Comparison between MaSA and retention in RMT-S's architecture.}
    \vspace{-4mm}
    \label{tab:pre_trp_cc}
\end{table}

\begin{table}[t]
    \centering
    \setlength{\tabcolsep}{3.8mm}
    \scalebox{0.75}{
    \begin{tabular}{c|c c c|c}
        \toprule[1pt]
        Model & \makecell{Params\\(M)} & \makecell{FLOPs$\downarrow$\\(G)} & \makecell{Throughput$\uparrow$\\(imgs/s)} & \makecell{Top1\\(\%)}\\
        \midrule[0.5pt]
        BiFormer-T~\cite{biformer} & 13 & 2.2 & 1602 &81.4 \\
        CMT-XS~\cite{cmt} & 15 & 1.5 & 1476 & 81.8 \\
        SMT-T~\cite{SMT} & 12 & 2.4 & 636 & 82.2 \\
        \rowcolor{gray!30}RMT-T & 14 & 2.5 & 1650 & 82.4 \\
        \midrule[0.5pt]
        % Swin-B & 88 & 15.5 & 756 & 83.5\\
        CMT-S~\cite{cmt} & 25 & 4.0 & 848 & 83.5 \\
        MaxViT-T~\cite{maxvit} & 31 & 5.6 & 826 & 83.6 \\
        SMT-S~\cite{SMT} & 20 & 4.8 & 356 & 83.7\\
        BiFormer-S~\cite{biformer} & 26 & 4.5 & 766 & 83.8\\
        \rowcolor{gray!30}RMT-Swin-T & 29 & 4.7 & 1192 & 83.6\\
        \rowcolor{gray!30}RMT-S & 27 & 4.5 & 876 & 84.1 \\
        \midrule[0.5pt]
        SMT-B~\cite{SMT} & 32 & 7.7 & 237 & 84.3 \\
        BiFormer-B~\cite{biformer} & 57 & 9.8 & 498 & 84.3\\
        CMT-B~\cite{cmt} & 46 & 9.3 & 447 & 84.5 \\
        MaxViT-S~\cite{maxvit} & 69 & 11.7 & 546 & 84.5 \\
        \rowcolor{gray!30}RMT-Swin-S & 50 & 9.1 & 722 & 84.5 \\
        \rowcolor{gray!30}RMT-B & 54 & 9.7 & 457 & 85.0 \\
        \midrule
        SMT-L~\cite{SMT} & 80 & 17.7 & 158 & 84.6 \\
        MaxViT-B~\cite{maxvit} & 120 & 23.4 & 306 & 84.9 \\
        \rowcolor{gray!30}RMT-L & 95 & 18.2 & 326 & 85.5\\
        \bottomrule[1pt]
    \end{tabular}}
    \caption{Comparison of inference speed among SOTA models.}
    \vspace{-5mm}
    \label{tab:eff1}
\end{table}

\paragraph{Results.}The results of semantic segmentation can be found in Tab.~\ref{tab:semanticfpn}. All the FLOPs are measured with the resolution of $512\times2048$, except the group of RMT-T, which are measured with the resolution of $512\times512$. All our models achieve the best performance in all comparisons. Specifically, our RMT-S exceeds Shunted-S for \textbf{+1.2} mIoU with Semantic FPN. Moreover, our RMT-B outperforms the recent InternImage-S for \textbf{+1.8} mIoU. All the above results demonstrate our model's superiority in dense prediction.

\subsection{Ablation Study}

\paragraph{Strict comparison with previous works.}In order to make a strict comparison with previous methods, we align RMT's hyperparameters (such as whether to use hierarchical structure, the number of channels in the four stages of the hierarchical model, whether to use positional encoding and convolution stem, etc.) of the overall architecture with DeiT~\cite{deit} and Swin~\cite{SwinTransformer}, and only replace the Self-Attention/Window Self-Attention with our MaSA. The comparison results are shown in Tab.~\ref{tab:ablation_study}, where RMT significantly outperforms DeiT-S, Swin-T, and Swin-S.

\vspace{-3mm}

\paragraph{MaSA.}We verify the impact of Manhattan Self-Attention on the model, as shown in the Tab.~\ref{tab:ablation_study}. MaSA improves the model's performance in image classification and downstream tasks by a large margin. Specifically, the classification accuracy of MaSA is \textbf{0.8\%} higher than that of vanilla attention.

\vspace{-3mm}

\paragraph{Softmax.}In RetNet, Softmax is replaced with a non-linear gating function to accommodate its various computational forms~\cite{retnet}. We replace the Softmax in MaSA with this gating function. However, the model utilizing the gating function cannot undergo stable training. It is worth noting that this does not mean the gating function is inferior to Softmax. The gating function may just not be compatible with our decomposed form or spatial decay.

\vspace{-3mm}

\paragraph{LCE.}Local Context Enhancement also plays a role in the excellent performance of our model. LCE improves the classification accuracy of RMT by 0.3\% and enhances the model's performance in downstream tasks.

\vspace{-3mm}

\paragraph{CPE.}Just like previous methods, CPE provides our model with flexible position encoding and more positional information, contributing to the improvement in the model's performance in image classification and downstream tasks. 

\vspace{-3mm}

\paragraph{Convolutional Stem.}The initial convolutional stem of the model provides better local information, thereby further enhancing the model's performance on various tasks. 

\vspace{-3mm}

\paragraph{Decomposed MaSA.}In RMT-S, we substitute the decomposed MaSA (MaSA-d) in the third stage with the original MaSA to validate the effectiveness of our decomposition method, as illustrated in Tab.~\ref{tab:ab_decop}. In terms of image classification, MaSA-d and MaSA achieve comparable accuracy. However, for semantic segmentation, employing MaSA-d significantly reduces computational burden while yielding similar result.

\vspace{-3mm}

\paragraph{MaSA v.s. Retention.}As shown in Tab.~\ref{tab:pre_trp_cc}, we replace MaSA with the original retention in the architecture of RMT-S. We partition the tokens into chunks using the method employed in Swin-Transformer~\cite{SwinTransformer} for chunk-wise retention. Due to the limitation of retention in modeling one-dimensional causal data, the performance of the vision backbone based on it falls behind RMT. Moreover, the chunk-wise and recurrent forms of retention disrupt the parallelism of the vision backbone, resulting in lower inference speed.

\vspace{-3mm}

\paragraph{Inference Speed.}We compare the RMT's inference speed with the recent best performing vision backbones in Tab.~\ref{tab:eff1}. Our RMT demonstrates the optimal trade-off between speed and accuracy.

\section{Conclusion}

In this work, we propose RMT, a vision backbone with explicit spatial prior. RMT extends the temporal decay used for causal modeling in NLP to the spatial level and introduces a spatial decay matrix based on the Manhattan distance. The matrix incorporates explicit spatial prior into the Self-Attention. Additionally, RMT utilizes a Self-Attention decomposition form that can sparsely model global information without disrupting the spatial decay matrix. The combination of spatial decay matrix and attention decomposition form enables RMT to possess explicit spatial prior and linear complexity. Extensive experiments in image classification, object detection, instance segmentation, and semantic segmentation validate the superiority of RMT.
% \newpage

% \clearpage
% \setcounter{page}{1}
% \maketitlesupplementary

\begin{appendices}

\section{Architecture Details}

\begin{table*}[ht]
    \centering
    \begin{tabular}{c|c c c c|c c}
         \toprule[1pt]
         Model & Blocks & Channels & Heads & Ratios & Params(M) & FLOPs(G)\\
         \midrule[0.5pt]
         RMT-T & [2, 2, 8, 2] & [64, 128, 256, 512] & [4, 4, 8, 16] & [3, 3, 3, 3] & 14 & 2.5 \\
         RMT-S & [3, 4, 18, 4] & [64, 128, 256, 512] & [4, 4, 8, 16] & [4, 4, 3, 3] & 27 & 4.5 \\
         RMT-B & [4, 8, 25, 8] & [80, 160, 320, 512] & [5, 5, 10, 16] & [4, 4, 3, 3] & 54 & 9.7 \\
         RMT-L & [4, 8, 25, 8] & [112, 224, 448, 640] & [7, 7, 14, 20] & [4, 4, 3, 3] & 95 & 18.2 \\
         \midrule
         RMT-DeiT-S & [12] & [384] & [6] & [4] & 22 & 4.6 \\
         \midrule[0.5pt]
         RMT-Swin-T & [2, 2, 6, 2] & [96, 192, 384, 768] & [3, 6, 12, 24] & [4, 4, 4, 4] & 29 & 4.7 \\
         RMT-Swin-S & [2, 2, 18, 2] & [96, 192, 384, 768] & [3, 6, 12, 24] & [4, 4, 4, 4] & 50 & 9.1 \\
         \bottomrule[1pt]
    \end{tabular}
    \caption{Detailed Architectures of our models.}
    \label{tab:arc}
\end{table*}

Our architectures are illustrated in the Tab.~\ref{tab:arc}. For convolution stem, we apply five $3\times3$ convolutions to embed the image into $56\times56$ tokens. GELU and batch normalization are used after each convolution except the last one, which is only followed by batch normalization. $3\times3$ convolutions with stride 2 are used between stages to reduce the feature map's resolution. $3\times3$ depth-wise convolutions are adopted in CPE. Moreover, $5\times 5$ depth-wise convolutions are adopted in LCE. RMT-DeiT-S, RMT-Swin-T, and RMT-Swin-S are models that we used in our ablation experiments. Their structures closely align with the structure of DeiT~\cite{deit} and Swin-Transformer~\cite{SwinTransformer} without using techniques like convolution stem, CPE, and others.

\section{Experimental Settings}
\paragraph{ImageNet Image Classification.}We adopt the same training strategy with DeiT~\cite{deit} with the only supervision is the classification loss. In particular, our models are trained from scratch for 300 epochs. We use the AdamW optimizer with a cosine decay learning rate scheduler and 5 epochs of linear warm-up. The initial learning rate, weight decay, and batch size are set to 0.001, 0.05, and 1024, respectively. Our augmentation settings are RandAugment~\cite{randomaugment} (randm9-mstd0.5-inc1), Mixup~\cite{mixup} (prob=0.8), CutMix~\cite{cutmix} (probe=1.0), Random Erasing~\cite{randera} (prob=0.25) and Exponential Moving Average (EMA)~\cite{EMA}. The maximum rates of increasing stochastic depth~\cite{droppath} are set to 0.1/0.15/0.4/0.5 for RMT-T/S/B/L, respectively. For a more comprehensive comparison, we train two versions of the model. The first version uses only classification loss as the supervision, while the second version, in addition to the classification loss, incorporates token labeling introduced by \cite{tokenlabel} for additional supervision. Models using token labeling are marked with``*".

\paragraph{COCO Object Detection and Instance Segmentation.}We apply RetinaNet~\cite{retinanet}, Mask-RCNN~\cite{maskrcnn} and Cascaded Mask-CNN~\cite{cai18cascadercnn} as the detection frameworks to conduct experiments. We implement them based on the MMDetection~\cite{mmdetection}. All models are trained under two common settings:``$1\times$" (12 epochs for training) and``$3\times$+MS" (36 epochs with multi-scale augmentation for training). For the ``$1\times$" setting, images are resized to the shorter side of 800 pixels. For the ``$3\times$+MS", we use the multi-scale training strategy and randomly resize the shorter side between 480 to 800 pixels. We apply AdamW optimizer with the initial learning rate of 1e-4. For RetinaNet, we use the weight decay of 1e-4 for RetinaNet while we set it to 5e-2 for Mask-RCNN and Cascaded Mask-RCNN. For all settings, we use the batch size of 16, which follows the previous works~\cite{SwinTransformer, focal, focalnet}

\paragraph{ADE20K Semantic Segmentation.}Based on MMSegmentation~\cite{mmsegmentation}, we implement UperNet~\cite{upernet} and SemanticFPN~\cite{semanticfpn} to validate our models. For UperNet, we follow the previous setting of Swin-Transformer~\cite{SwinTransformer} and train the model for 160k iterations with the input size of $512\times 512$. For SemanticFPN, we also use the input resolution of $512\times512$ but train the models for 80k iterations.

\section{Efficiency Comparison}

We compare the inference speed of RMT with other backbones, as shown in Tab.~\ref{tab:eff2}. Our models achieve the best trade-off between speed and accuracy among many competitors.

\begin{table*}[t]
    \centering
    \subfloat{
    \begin{tabular}{c|c c c|c}
    \toprule[1pt]
         Model & \makecell{Params\\(M)} & \makecell{FLOPs\\(G)} & \makecell{Troughput\\(imgs/s)} & \makecell{Top1\\(\%)}\\
         \midrule
         MPViT-XS~\cite{mpvit} & 11 & 2.9 & 1496 & 80.9 \\
         Swin-T~\cite{SwinTransformer} & 29 & 4.5 & 1704 & 81.3 \\
         BiFormer-T~\cite{biformer} & 13 & 2.2 & 1602 &81.4 \\
         GC-ViT-XT~\cite{globalvit} & 20 & 2.6 & 1308 & 82.0 \\
         SMT-T~\cite{SMT} & 12 & 2.4 & 636 & 82.2 \\
         \rowcolor{gray!30}RMT-T & 14 & 2.5 & 1650 & 82.4 \\
         \midrule
         Focal-T~\cite{focal} & 29 & 4.9 & 582 & 82.2 \\
         CSWin-T~\cite{cswin} & 22 & 4.3 & 1561 & 82.7\\
         Eff-B4~\cite{efficientnet} & 19 & 4.2 & 627 & 82.9 \\
         MPViT-S~\cite{mpvit} & 23 & 4.7 & 986 & 83.0 \\
         Swin-S~\cite{SwinTransformer} & 50 & 8.8 & 1006 & 83.0 \\
         SGFormer-S~\cite{sgformer} & 23 & 4.8 & 952 & 83.2 \\
         iFormer-S~\cite{iformer} & 20 & 4.8 & 1051 & 83.4 \\
         CMT-S~\cite{cmt} & 25 & 4.0 & 848 & 83.5 \\
         \rowcolor{gray!30}RMT-Swin-T & 29 & 4.7 & 1192 & 83.6\\
         CSwin-S~\cite{cswin} & 35 & 6.9 & 972 & 83.6 \\
         MaxViT-T~\cite{maxvit} & 31 & 5.6 & 826 & 83.6 \\
         SMT-S~\cite{SMT} & 20 & 4.8 & 356 & 83.7\\
         BiFormer-S~\cite{biformer} & 26 & 4.5 & 766 & 83.8\\
         \rowcolor{gray!30}RMT-S & 27 & 4.5 & 876 & 84.1 \\
         \bottomrule[1pt]
    \end{tabular}}
    \subfloat{
    \begin{tabular}{c|c c c|c}
    \toprule[1pt]
         Model & \makecell{Params\\(M)} & \makecell{FLOPs\\(G)} & \makecell{Troughput\\(imgs/s)} & \makecell{Top1\\(\%)}\\
         \midrule
         Focal-S~\cite{focal} & 51 & 9.1 & 351 & 83.5 \\
         Eff-B5~\cite{efficientnet} & 30 & 9.9 & 302 & 83.6 \\
         SGFormer-M~\cite{sgformer} & 39 & 7.5 & 598 & 84.1 \\
         SMT-B~\cite{SMT} & 32 & 7.7 & 237 & 84.3 \\
         BiFormer-B~\cite{biformer} & 57 & 9.8 & 498 & 84.3\\
         \rowcolor{gray!30}RMT-Swin-S & 50 & 9.1 & 722 & 84.5 \\
         MaxViT-S~\cite{maxvit} & 69 & 11.7 & 546 & 84.5 \\
         CMT-B~\cite{cmt} & 46 & 9.3 & 447 & 84.5 \\
         iFormer-B~\cite{iformer} & 48 & 9.4 & 688 & 84.6 \\
         \rowcolor{gray!30}RMT-B & 54 & 9.7 & 457 & 85.0 \\
         \midrule
         Swin-B~\cite{SwinTransformer} & 88 & 15.5 & 756 & 83.5\\
         Eff-B6~\cite{efficientnet} & 43 & 19.0 & 172 & 84.0 \\
         Focal-B~\cite{focal} & 90 & 16.4 & 256 & 84.0 \\
         CSWin-B~\cite{cswin} & 78 & 15.0 & 660 & 84.2 \\
         MPViT-B~\cite{mpvit} & 75 & 16.4 & 498 & 84.3 \\
         SMT-L~\cite{SMT} & 80 & 17.7 & 158 & 84.6 \\
         SGFormer-B~\cite{sgformer} & 78 & 15.6 & 388 & 84.7 \\
         iFormer-L~\cite{iformer} & 87 & 14.0 & 410 & 84.8 \\
         MaxViT-B~\cite{maxvit} & 120 & 23.4 & 306 & 84.9 \\
         \rowcolor{gray!30}RMT-L & 95 & 18.2 & 326 & 85.5 \\
         \bottomrule[1pt]
    \end{tabular}}
    \caption{Comparison of inference speed.}
    \label{tab:eff2}
\end{table*}

\section{Details of Explicit Decay}

We use different $\gamma$ for each head of the multi-head ReSA to control the receptive field of each head, enabling the ReSA to perceive multi-scale information. We keep all the $\gamma$ of ReSA's heads within a certain range. Assuming the given receptive field control interval of a specific ReSA module is $[a, b]$, where both $a$ and $b$ are positive real numbers. And the total number of the ReSA module's heads is $N$. The $\gamma$ for its $i$th head can be written as Eq.~\ref{eq:gamma}:
\begin{equation}
    \label{eq:gamma}
    \gamma_i = 1-2^{-a-\frac{(b-a)i}{N}}
\end{equation}

For different stages of different backbones, we use different values of $a$ and $b$, with the details shown in Tab.~\ref{tab:range}. 

\begin{table}[t]
    \centering
    \begin{tabular}{c|c c}
    \toprule[1pt]
         Model & $a$ & $b$ \\
         \midrule[0.5pt]
         RMT-T & [2, 2, 2, 2] & [6, 6, 8, 8] \\
         RMT-S & [2, 2, 2, 2] & [6, 6, 8, 8] \\
         RMT-B & [2, 2, 2, 2] & [7, 7, 8, 8] \\
         RMT-L & [2, 2, 2, 2] & [8, 8, 8, 8] \\
         \midrule
         RMT-DeiT-S & [2] & [8]\\
         \midrule
         RMT-Swin-T & [2, 2, 2, 2] & [8, 8, 8, 8] \\
         RMT-Swin-S & [2, 2, 2, 2] & [8, 8, 8, 8] \\
    \bottomrule[1pt]
    \end{tabular}
    \caption{Details about the $\gamma$ decay.}
    \label{tab:range}
\end{table}

\end{appendices}

\newpage
{
    \small
    \bibliographystyle{ieeenat_fullname}
    \bibliography{main}
}


\end{document}